\documentclass{article}
\usepackage{iclr2027_conference,times}

\usepackage{amsmath,amsfonts,bm}

\def\eqref#1{equation~\ref{#1}}

\def\1{\bm{1}}

\def\va{{\bm{a}}}
\def\vb{{\bm{b}}}

\def\vx{{\bm{x}}}

\DeclareMathAlphabet{\mathsfit}{\encodingdefault}{\sfdefault}{m}{sl}
\SetMathAlphabet{\mathsfit}{bold}{\encodingdefault}{\sfdefault}{bx}{n}

\DeclareMathOperator*{\argmin}{arg\,min}

\usepackage{hyperref}
\usepackage{url}
\hypersetup{
  hidelinks,
  pdftitle={Coupled Harness--Model Evolution for Scientific Equation Discovery},
  pdfauthor={Pengze Li, Jiaquan Zhang, Encheng Su, Yunbo Long, Zihang Zeng, Xinping Liu, Han Zhang, Junchi Yu, Shixiang Tang, Aoran Wang, Xi Chen}
}
\usepackage{amsmath}
\usepackage{amssymb}
\usepackage{booktabs}
\usepackage{graphicx}
\usepackage{wrapfig}
\usepackage{algorithm}
\usepackage{algpseudocode}

\title{Coupled Harness--Model Evolution \\ for Scientific Equation Discovery}

\author{\parbox{\dimexpr\textwidth-2\tabcolsep\relax}{\centering\small
\textbf{Pengze Li$^{1, 7}$, Jiaquan Zhang$^{1}$, Encheng Su$^{2}$, Yunbo Long$^{3}$,}\\
\textbf{Zihang Zeng$^{1}$, Xinping Liu$^{1}$, Han Zhang$^{4}$, Junchi Yu$^{5}$,}\\
\textbf{Shixiang Tang$^{6}$, Aoran Wang$^{7}$, Xi Chen$^{1}$}\\[4pt]
\normalfont\small
$^{1}$Fudan University\\
$^{2}$University of Science and Technology of China\\
$^{3}$University of Cambridge\quad $^{4}$Peng Cheng Laboratory\\
$^{5}$University of Oxford\quad $^{6}$The Chinese University of Hong Kong\\
$^{7}$Shanghai Artificial Intelligence Laboratory
}}

\iclrfinalcopy
\begin{document}

\maketitle
\pagestyle{plain}
\thispagestyle{plain}

\begin{abstract}
Large language models (LLMs) are increasingly used to discover scientific equations from
experimental data. Such a discovery system has two parts that must work together: a library of
executable computational tools and the model's tool-use policy. The tools supply the solving capability, from diagnosing
structure in the data to fitting candidate equations; the policy plans and orchestrates their use. Yet existing systems adapt only one
part, keeping either the toolset or the model fixed, which leaves the corresponding failures
uncorrectable. We develop a coupled evolution framework
that improves both parts from the same verifiable fit-error signal. In the tool-evolution phase, an agent grows an executable tool library
from a small set of generic numerical operations by diagnosing its own failures: it authors a new
tool when a capability is missing and revises its selection rules when the right tool went unused.
In the policy-evolution phase, a small open model (roughly 3B active parameters) is fine-tuned on
successful tool-use trajectories collected with the frozen library, on training formulas screened
to be disjoint from all evaluation benchmarks. Ablations show the two phases are complementary:
with no tool library at all the same model solves almost nothing, the tools account for most of the
gain, and trajectory training adds a further gain that is largest on the most difficult problems. The full system solves 79.1/71.3\% of LSR-Synth problems at fit tolerances
$10^{-2}$/$10^{-3}$, reaches 63.8\% pooled accuracy on the distributionally different SRSD-Feynman
panel, and obtains a normalized CP3-28 aggregate of $-7.27$. These results suggest a broader recipe
for scientific agents: let the model and its instruments improve each other under verifiable
feedback.
\end{abstract}

\section{Introduction}
\label{sec:intro}

Many scientific laws are written as equations relating measured quantities. Given experimental
observations, scientific equation discovery, classically studied as symbolic regression (SR), asks
for a compact equation that fits the observed data and remains accurate on new observations; it
has long been a central goal of automated science \citep{brunton2016sindy,cranmer2023pysr}.
Classical methods search the space of candidate expressions directly, by genetic programming or
neural-guided search \citep{cranmer2023pysr,petersen2021dsr,landajuela2022udsr}. Recent systems
instead place a large language model (LLM) at the center of the search: the model proposes
candidate equations, calls computational tools, and reads the returned numbers to decide what to
try next \citep{shojaee2025llmsr,grayeli2024lasr,merler2024icsr,xia2026srscientist}. These tools
are executable numerical routines that, for example, detect periodicity in the data, fit the
unknown constants of a proposed equation, or score how well a candidate matches the observations.

Two parts of such a system jointly determine what it can discover. The first is the tool library,
the set of executable operations the model can call. The library bounds the reachable hypothesis
space: an equation family that no available tool can fit or test can never be confirmed, no matter
what the model proposes. The second is the model's tool-use policy, its learned behavior of
choosing which tool to call, with what arguments, and how to act on the result. Even when the right
tool exists, a policy that skips it, calls it with unsuitable arguments, or misreads its output
still fails. A failed discovery attempt therefore has two distinct causes, a missing capability or
a misused one, and correcting them requires changing different parts of the system: the first
requires new tools, the second a better policy.

Existing approaches, however, typically adapt only one of the two parts. One line holds the model
fixed and grows the tools or skills around it, in general agent settings
\citep{wang2024voyager,qian2023creator,yuan2024craft,wang2024trove} as well as in equation
discovery, where systems such as SR-Scientist keep the model frozen while it writes analysis code
and refines candidate equations \citep{xia2026srscientist,shojaee2025llmsr,yang2026kepleragent}. A
complementary line trains the model while holding its toolset fixed, from general tool-use training
\citep{schick2023toolformer,patil2024gorilla,qin2024toollm} to SR systems that fine-tune the model
on discovery trajectories with an unchanged toolset \citep{zhao2026asr,behzadifar2026decaevolve}.
Either choice leaves one class of failures uncorrectable, and to our knowledge no equation-discovery
system evolves its tool library and trains its tool-use policy together
(Section~\ref{sec:related}). Equation discovery is also a
setting where this coupling matters and can be studied cleanly: the equation families a new problem
requires are hard to anticipate, so a library fixed at design time risks lacking a needed
capability; whether a tool helps depends on the policy that invokes it, so added capability pays
off only when the model learns to use it; and every attempt ends with a verifiable outcome, the fit
error of the proposed equation on observations the system has not seen, which separates success
from failure unambiguously and can drive the improvement of both parts.

We therefore evolve both parts under the same verifiable fit-error signal. Throughout the paper we
call everything outside the model the \emph{harness}: the executable tool library, together with
the rules that present the tools to the model and the rule that selects the final candidate
equation. Our framework proceeds in two phases, each updating one part while the other is held
fixed. Phase~A evolves the harness: after each failed attempt, the system diagnoses whether the
failure came from a missing capability or from a wrong choice by the model, and accordingly writes
a new tool or revises the presentation and selection rules. Phase~B improves the model's tool-use
policy: with the evolved harness frozen, a small open model is fine-tuned on successful tool-use
trajectories. In this way, failures expand what the system can test, and successes deepen the model's
command of its tools, teaching it when to invoke which tool, how to set its arguments, and how to
act on the returned evidence in the search.

This paper makes three contributions.
\begin{itemize}
  \item We identify a coupled bottleneck in LLM-based equation discovery: a failed attempt can
  stem from a missing tool capability or from a misused one, and adapting only the toolset or only
  the model leaves one of the two failure classes uncorrectable.

  \item We propose, to our knowledge, the first framework that resolves both bottlenecks in
  scientific equation discovery, coupling failure-driven tool evolution with trajectory-based
  policy training under one verifiable fit-error signal.

  \item We provide evidence that the evolved capability transfers across benchmarks: the tool
  library and the trained policy, both produced on one benchmark family, are applied unchanged to benchmarks with different equation distributions and surpass
  the best published results there, while matched ablations attribute most of the gain to the
  evolved library and a further gain in tool-use reliability to policy training.
\end{itemize}

\section{Related Work}
\label{sec:related}

Prior work relevant to this paper falls into four lines: evolving tools around a frozen model,
training a model to use a fixed toolset, co-evolving a model with its capability assets, and
symbolic regression itself.

\paragraph{Tool evolution around a frozen model.}
The first line improves an agent by evolving its scaffolding while keeping the model fixed.
Voyager \citep{wang2024voyager} accumulates a reusable skill library while exploring an open-ended
environment, and a line of tool-making methods including CREATOR \citep{qian2023creator}, CRAFT
\citep{yuan2024craft}, TroVE \citep{wang2024trove}, and LLMs-as-Tool-Makers
\citep{cai2024latm} lets a model author, curate, and reuse its own tools. Automated design of
agentic systems \citep{hu2025adas} and evolutionary program search \citep{novikov2025alphaevolve}
push further, searching over the control flow and code of the agent itself. EVOTOOL
\citep{yang2026evotool} and ToolMaker \citep{wolflein2025toolmaker} evolve tool-use policies and
create agent tools automatically. All of these hold the model fixed.

\paragraph{Tool-use training on a fixed toolset.}
A complementary line does the opposite, teaching the model to call a fixed set of tools or APIs:
through self-supervised demonstrations \citep{schick2023toolformer}, fine-tuning for API calls
\citep{patil2024gorilla}, large-scale tool-use instruction tuning \citep{qin2024toollm}, and
reward-driven tool learning \citep{qian2025toolrl}; see \citet{qu2024toolsurvey} for a survey. Here
the tool set is designed by humans and held constant while the model learns.

\paragraph{Model--asset co-evolution.}
A recent frontier co-evolves both objects in a single loop. Skill Self-Play
\citep{huang2026skillselfplay} optimizes proposer/solver policies by self-play RL while a controller
induces, rewrites, and prunes a skill library; Mem$^2$Evolve \citep{cheng2026mem2evolve} jointly
evolves experience memory and dynamic tools/expert agents; and further variants co-evolve a policy
with tasks \citep{yang2026coevolve}, training harnesses \citep{chen2026evotrainer}, environments
\citep{hu2026seal}, or world models \citep{fang2025webevolver}. These works operate in open-ended or
game-like settings, where progress is scored by self-play or proxy signals.

\paragraph{Symbolic regression.}
Within symbolic regression itself, classical methods search directly over expression space, via
genetic programming as in PySR \citep{cranmer2023pysr}, neural-guided search as in DSR/uDSR
\citep{petersen2021dsr,landajuela2022udsr}, and sparse identification of dynamics
\citep{brunton2016sindy}. A recent wave uses LLMs as the search driver: LLM-SR
\citep{shojaee2025llmsr} proposes and refines equation programs from feedback, LaSR
\citep{grayeli2024lasr} accelerates search with learned concept abstractions, in-context SR
\citep{merler2024icsr} and LLM equation discovery \citep{du2024llmed} prompt the model directly, and
SR-Scientist \citep{xia2026srscientist} casts the LLM as an autonomous agent that writes analysis code
and optimizes candidate equations against feedback. Closest in spirit to co-evolution within SR
are A-SR \citep{zhao2026asr}, which distills trajectories and evolves the search process around
candidate equations; DecAEvolve \citep{behzadifar2026decaevolve}, which combines test-time LoRA/GRPO
adaptation with evolution of candidate equation programs; and KeplerAgent
\citep{yang2026kepleragent}, which configures PySR/PySINDy while retaining the solver model.
LLM-SRBench \citep{shojaee2025llmsrbench} provides a standard benchmark suite whose
out-of-distribution design distinguishes discovery from interpolation.

\paragraph{Position of this work.}
Our approach differs from each line in one specific way. From the tool-evolution line we inherit
the idea of growing reusable tools, but add a failure-diagnosis step that assigns each failure to
either a missing capability or a wrong selection, and an acceptance check that makes accumulated
capability monotone. Unlike the tool-use training line, our training phase operates on an evolved
tool set rather than a human-designed one. Relative to model--asset co-evolution, equation
discovery is a harder test of the idea: the reward is a hard, externally verifiable fit criterion,
the artifacts must generalize out of distribution, and any gain must survive a memorization check;
these requirements drive the diagnosis, acceptance, and data-provenance machinery we introduce.
Within SR, SR-Scientist is our closest head-to-head, as it is agentic, tool-using, and evaluated on
LSR-Synth with the same metric; among the publicly available SR literature, however, we are not
aware of a prior system that couples failure-driven evolution of an executable SR tool library with
trajectory-based training of the equation-discovery model.

\section{Method}
\label{sec:method}

\subsection{Problem setup and overview}
\label{sec:method-setup}

\paragraph{Problem instances.}
A problem instance presents the system with a table of experimental observations. Each observation
pairs an input vector $\vx_i\in\mathbb R^d$, holding the measured input quantities (e.g.,
temperature and concentration in a chemistry problem), with a scalar outcome $y_i$; the goal is to
recover a closed-form equation $y\approx f(\vx)$ underlying the data. The observations are
partitioned into three disjoint subsets,
\begin{equation}
\mathcal P=(\mathcal D_{\rm fit},\mathcal D_{\rm sel},\mathcal D_{\rm test}),
\qquad
\mathcal D_z=\{(\vx_i,y_i)\}_{i\in I_z},\quad
z\in\{\mathrm{fit},\mathrm{sel},\mathrm{test}\},
\end{equation}
where $I_z$ are the corresponding index sets. The three subsets play different roles. The
\emph{fitting subset} $\mathcal D_{\rm fit}$ is the only data the system sees while it searches for
equations; the unknown constants of every candidate equation are also fitted on it. The
\emph{selection subset} $\mathcal D_{\rm sel}$ is a validation set: it is read only after the
search ends, to choose among the candidate equations the search produced. The \emph{test subset}
$\mathcal D_{\rm test}$ is withheld from the system entirely and is read only by the external
scorer that grades the final equation. We call
$\mathcal D_{\rm pub}=\mathcal D_{\rm fit}\cup\mathcal D_{\rm sel}$ the \emph{public} data, since
the system may eventually see both parts, in contrast to the withheld test subset. Standard SR
benchmarks specify only fitting and test data; the selection subset is a validation split we carve
out of the public data for this study, not part of the original problem definition.

A \emph{candidate equation} $c$ is an expression tree, that is, a symbolic expression such as
$c(\vx)=\theta_1 e^{-\theta_2 x_1}\sin(\theta_3 x_2)$, whose numeric constants $\theta$ are fitted
on $\mathcal D_{\rm fit}$. Given the tolerance $\tau$ specified by a benchmark, a candidate counts
as solving the instance if its relative prediction error stays within $\tau$ on every test
observation:
\begin{equation}
\label{eq:solve}
\mathrm{solve}_\tau(c,\mathcal P)=\mathbf 1\!\left[
\max_{i\in I_{\rm test}}\;
\frac{\lvert c(\vx_i)-y_i\rvert}{\max(\lvert y_i\rvert,10^{-12})}\le\tau\right].
\end{equation}
Because the error is normalized by the target's magnitude, the criterion is dimensionless and one
tolerance applies across problems regardless of the outcome's units or scale; the constant
$10^{-12}$ guards near-zero targets. The task is therefore a coupled discrete--continuous search:
propose expression structures (discrete) and fit their constants (continuous), using only the
public observations.

\paragraph{System and notation.}
The system pairs an LLM with the harness introduced in Section~\ref{sec:intro}. Formally, the
harness is a triple $\mathcal H=(\mathcal L,\rho,\sigma)$: $\mathcal L$ is the executable tool
library; $\rho$ is the \emph{interaction interface}, the rules that describe the available tools to
the model and format tool results back to it; and $\sigma$ is the \emph{selection rule} that picks
the final equation among candidates. The model interacts with the harness in discrete steps. At
step $t$, the model observes a state $s_t$, which collects everything visible to it so far: the
problem's variable names, a preview of the fitting data, and the results of all previous tool
calls. It then emits an action $a_t$, either a call to a tool in $\mathcal L$ with chosen
arguments, or a new candidate equation. The harness executes the action on the fitting data,
appends the numerical result to the state, and adds any new candidate to the \emph{candidate
archive} $\mathcal C_t$, the set of candidate equations produced so far:
\begin{equation}
s_{t+1},\mathcal C_{t+1}=\mathsf H_{\mathcal H}(s_t,a_t;\mathcal D_{\rm fit}),\qquad
a_t\sim\pi(\cdot\mid s_t),
\end{equation}
where $\mathsf H_{\mathcal H}$ denotes one such interaction step executed by the harness configured
as $\mathcal H$. The model's \emph{tool-use policy} $\pi$ is its conditional distribution over
actions given the state; a \emph{trajectory} is the complete sequence of states and actions
produced on one problem instance.

\begin{figure*}[t]
    \centering
    \includegraphics[width=\textwidth]{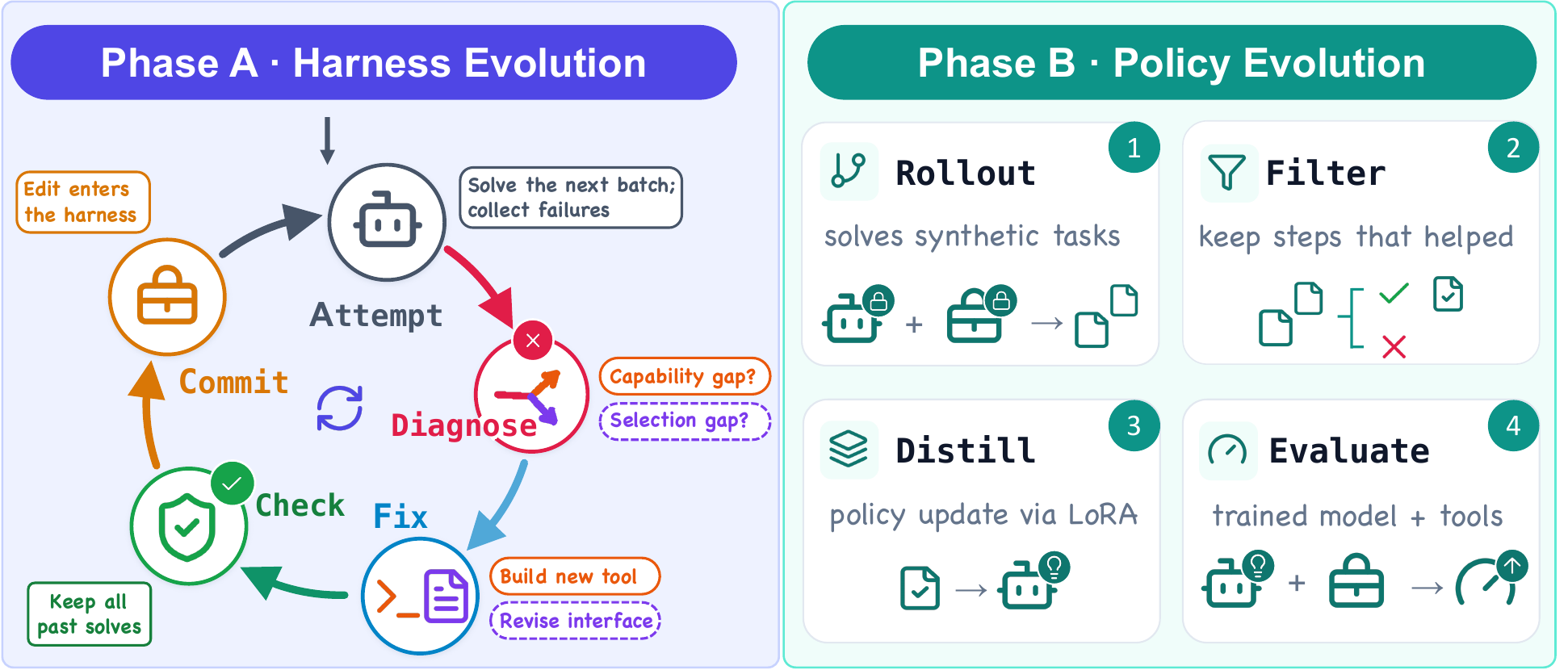}
    \caption{Overview of the coupled evolution. Phase A (left) starts from a generic tool set and evolves the harness by diagnosing failed attempts: a missing capability adds a tool, a wrong choice revises the interaction or selection rules. Phase B (right) holds the frozen harness fixed and fine-tunes an open Qwen controller on successful tool-use trajectories, yielding the reported model~+~tools system.}
    \label{fig:overview}
\end{figure*}

The coupled procedure applies two updates to these objects in sequence
(Figure~\ref{fig:overview}). Phase~A holds the model weights
fixed and updates the harness $\mathcal H$; Phase~B holds the harness fixed and updates the policy
$\pi$. This separation makes the two contributions identifiable in the ablations. Throughout both
phases, the information boundary above is preserved: during a trajectory, the model and harness
read only $\mathcal D_{\rm fit}$; after the trajectory ends, the selection rule $\sigma$ may read
$\mathcal D_{\rm sel}$; and only the external scorer reads $\mathcal D_{\rm test}$.

\subsection{Phase A: failure-driven harness evolution}
\label{sec:method-evolve}
Phase~A holds the model weights fixed and updates only the harness $\mathcal H$. Its input is a
\emph{development stream}: a sequence of problem instances drawn from the development benchmarks
(Section~\ref{sec:exp-setup}), partitioned into batches $B_1,\ldots,B_T$ that are processed one at
a time. Throughout the stream, the loop keeps track of the set of instances the current harness has
already solved.

\paragraph{Trajectory generation.}
During Phase~A the acting model is a frontier model, GPT-5.5~\citep{openai2026gpt55}, accessed
through the OpenAI API. It works through each instance in the current batch with the current
harness, producing a trajectory and a final equation per instance. The initial library
$\mathcal L_0$ contains 25 generic numerical tools (e.g., axis transforms, residual refits, and
basic verification). If every instance in the batch is solved, the harness is left unchanged and
the loop advances to the next batch; otherwise, each unsolved trajectory is passed to failure
classification.

\paragraph{Failure classification.}
A second instance of GPT-5.5, acting as the \emph{reflection model}, reads each unsolved trajectory
and diagnoses it as one of two failure types. A \emph{capability gap} means the library lacks an
operation the instance needed, for example a fitter for the relevant equation family; the
reflection model then authors a new tool, or revises an existing one, in restricted Python. A
\emph{selection gap} means a suitable tool existed but the model skipped it, called it with
unsuitable arguments, or misread its output; the reflection model then revises the interaction
interface $\rho$ or the selection rule $\sigma$ instead. The reflection model operates only during
Phase~A; it produces no Phase-B training data and is absent at evaluation.

\paragraph{Monotone acceptance.}
Each proposed edit is provisional until it passes two checks. First, a newly authored tool must
pass a \emph{promotion gate}: it is run on data generated for this purpose during authoring and
held out from the tool's construction, and is admitted only if it behaves as specified there.
Second, the loop re-runs the revised harness on all previously solved instances and accepts the
edit only if every one of them remains solved, so accumulated capability never regresses. Accepted
edits are committed before the loop advances to the next batch. After the last batch, the resulting
harness is frozen; we denote it $\mathcal H^\star$ and use it unchanged in Phase~B and at
evaluation. The frozen toolkit and its provenance analysis are summarized in
Appendices~\ref{app:taxonomy} and~\ref{app:tool-check}, and Algorithm~\ref{alg:evolve} in
Appendix~\ref{app:loop} states the full loop.

\subsection{Phase B: policy evolution by trajectory distillation}
\label{sec:method-coevolve}
Phase~B holds the frozen harness $\mathcal H^\star$ fixed and trains a small open model to drive
it. The trained model is Qwen3.6-35B-A3B ($\sim$3B active parameters); we call it the
\emph{controller}, in contrast to the frontier model of Phase~A. Training updates the controller's
tool-use policy $\pi_\theta$: from each state $s_t$, the policy chooses the next tool call or
candidate equation. Because the training problems are synthesized and screened against every
evaluation benchmark (below), the update can improve only how the controller uses the tools.

\paragraph{Programmatic task synthesis.}
Two programmatic generators
produce ground-truth formulas featuring multi-variable cross-interactions and nested special
functions of the form $f(g(\cdot))$, and each formula is instantiated as a problem tuple
$\mathcal P$ with the same fit/selection/test structure used at evaluation.

\paragraph{Overlap firewall and information boundary.}
Let $\mathcal E$ be the set of ground-truth formulas of all evaluation benchmarks. A synthesized
formula $c$ enters the training pool only if $c\not\sim e$ for every $e\in\mathcal E$, where $\sim$
holds if the two formulas match under any of three equivalence tests: canonical symbolic equality,
affine reparametrization, or agreement on a bank of numeric probe points
(Appendix~\ref{app:firewall}). On the admitted instances, GPT-5.5 acts as the teacher: it solves
each instance with the frozen harness under the same information boundary used at evaluation, and
the recorded trajectories yield state/action pairs $\mathcal T=\{(s_t,a_t)\}$.

\paragraph{Training setup.}
The controller is fine-tuned on the subset of teacher steps that demonstrably helped:
\begin{equation}
\theta^\star=\argmin_\theta -\sum_{(s,a)\in\mathcal T_+}
\log \pi_\theta(a\mid s),\qquad
\mathcal T_+=\{(s_t,a_t)\in\mathcal T:\chi(s_t,a_t)=1\},
\end{equation}
where the filter $\chi$ keeps a step if its action produced a candidate that was promoted, improved
the fit error on the public data, or repaired a malformed tool call. This is cross-model
distillation: the teacher trajectories come from GPT-5.5, and the separately initialized Qwen
controller is the student, trained as a LoRA adapter on the retained action tokens. The complete
recipe is given in Appendix~\ref{app:recipe}; extensions that did not improve on this recipe,
including large-scale reinforcement learning, are recorded in Appendix~\ref{app:failures}.

\paragraph{Candidate selection.}
After a trajectory ends, the selection rule $\sigma$ picks the single equation to submit from the
candidate archive $\mathcal C$; the test subset never enters this choice. On benchmarks scored by
accuracy at a tolerance (Section~\ref{sec:exp-setup}), $\sigma$ is the Bayesian information
criterion computed on the public data,
\begin{equation}
\label{eq:bic}
\hat{c} \;=\; \argmin_{c\,\in\,\mathcal{C}} \; n \log \widehat{\mathrm{MSE}}_{\mathrm{pub}}(c) \;+\; k(c) \log n,
\end{equation}
where $\widehat{\mathrm{MSE}}_{\mathrm{pub}}(c)$ is the mean squared error of $c$ over the
$n=|\mathcal D_{\rm pub}|$ public observations and $k(c)$ is the size of the expression tree,
so that among comparably fitting candidates the simpler equation wins. On benchmarks scored by
regression error, $\sigma$ instead minimizes the error on the selection subset plus a light
complexity term. The same rule is applied to every system variant compared in the experiments;
details are in Appendix~\ref{app:selection}.

\section{Experiments}
\label{sec:experiments}
We evaluate the system across four equation-discovery panels, use matched ablations to isolate the
contributions of the evolved library and trajectory distillation, and analyze capability growth over
the evolution trace.

\subsection{Benchmarks and metrics}
\label{sec:exp-setup}
\paragraph{Benchmarks.}
Our primary benchmark is the LSR-Synth subset of LLM-SRBench \citep{shojaee2025llmsrbench}: 129
synthetic-yet-realistic discovery problems across chemistry, biology, physics, and materials science,
each with a public fitting split, a held-out in-domain test split, and an official
out-of-distribution (OOD) split with shifted input ranges. Three further panels widen the
comparison: LSR-Transform (111 problems recast from known physics, LSRT111) from the same family;
SRSD-Feynman-240 \citep{matsubara2022srsd}, 120 Feynman equations in three difficulty tiers, each
also posed with dummy variables that a solution must ignore; and CP3-28, 28 cosmology tasks with 12
published baseline methods \citep{thing2024cp3bench}. Within LLM-SRBench,
LSR-Synth and LSRT111 provide the Phase-A
development stream. We inspect the resulting library for benchmark-specific formula content
(Appendix~\ref{app:tool-check}), and screen Phase-B's synthetic formulas with a separate formula-level
overlap firewall (Appendix~\ref{app:firewall}). SRSD-Feynman and CP3-28 serve as
cross-benchmark tests, held out from both tool evolution and model training.

\paragraph{Metrics.}
Each panel is scored under its own native convention. LSR-Synth uses accuracy at two
fit-error tolerances, Acc@$10^{-2}$/Acc@$10^{-3}$, while LSRT111 uses Acc@$10^{-1}$/Acc@$10^{-2}$;
a problem counts as solved by the criterion of Equation~\ref{eq:solve}, i.e., the maximum relative
error over the test points must stay within the tolerance. Published baselines relax this by
discarding each problem's worst 5\% of test points before taking the maximum (the trimmed-5\%
convention); all our numbers, including the whole ablation table, keep every point (all-points),
which is strictly harder, and we quote trimmed-5\% only where a matched-convention comparison is
needed. SRSD-240 uses its original accuracy at $R^2>0.999$; CP3-28 publishes only per-problem MSE,
which we aggregate as the mean $\log_{10}$NMSE (Section~\ref{sec:exp-main}).

\paragraph{Systems compared.}
Our principal system pairs a Qwen3.6-35B-A3B controller (35B total, approximately 3B active
parameters) with the frozen evolved library; the controller is trained on formula-disjoint synthetic
trajectories (Section~\ref{sec:method-coevolve}). Tables label this condition model~+~tools.
Matched ablations use the untrained backbone with
the evolved library and a deterministic traversal of the library without a model. The traversal
calls each tool once, in a fixed order and with fixed default arguments, and submits from the
resulting candidates by the common selection rule
$\sigma$. It covers the 32 executable analysis tools of the catalog; the Python tool builder is
excluded, since authoring a new tool is itself a model action. The same traversal run on the initial
pre-evolution library isolates what Phase~A added. These conditions
are reported together in Section~\ref{sec:exp-compose} (Table~\ref{tab:ablation}); the untrained and
trained controllers isolate the training effect on a fixed toolkit. External model-based baselines retain
their published backbones: SR-Scientist
\citep{xia2026srscientist} on GPT-OSS-120B and Qwen3-Coder-480B; A-SR \citep{zhao2026asr}, fine-tuning
Llama3.1-8B; and LLM-SR \citep{shojaee2025llmsr} on Qwen3-Coder-480B,
plus non-LLM references \citep{cranmer2023pysr,petersen2021dsr} for context.

\subsection{Main results}
\label{sec:exp-main}

\begin{wrapfigure}{r}{0.50\linewidth}
\vspace{-6pt}
\includegraphics[width=\linewidth]{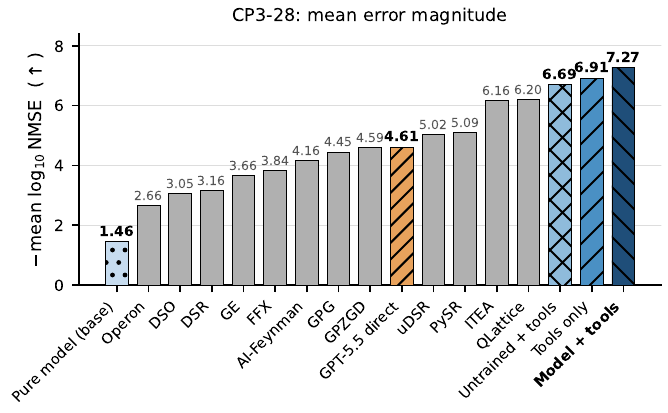}
\vspace{-4pt}
\caption{CP3-28 cosmology results over 28 problems. Bars show negated mean $\log_{10}$NMSE
(higher is better); GPT-5.5 direct is included as a tool-free reference. Appendix~\ref{app:cp3}
gives the aggregation procedure and per-problem comparisons.}
\label{fig:cp3_ode}
\vspace{-8pt}
\end{wrapfigure}

\begin{table*}[t]
\caption{LSR-Synth (129 problems) and LSR-Transform-111 accuracy (\%). All entries are percentages;
slash-separated cells report the first/second accuracy threshold. LSR-Synth cells use
Acc@$10^{-2}$/Acc@$10^{-3}$, while the LSRT111 column uses Acc@$10^{-1}$/Acc@$10^{-2}$. Bold marks the highest displayed
value for each benchmark and threshold within the reported conventions. ``--'' marks cells the source
does not report.}
\label{tab:lsr}
\begin{center}
\footnotesize
\setlength{\tabcolsep}{3pt}
\begin{tabular*}{\textwidth}{@{\extracolsep{\fill}}lcccccc@{}}
\toprule
{\bf System} & {\bf Aggregate} & {\bf Material} & {\bf Chemistry} & {\bf Biology} & {\bf Physics} & {\bf LSRT111} \\
\midrule
\multicolumn{7}{@{}l}{Traditional SR systems (worst-5\% trimmed):} \\
PySR & 29.5/14.5 & 53.3/22.7 & 25.9/11.1 & 16.7/7.0 & 25.8/16.7 & 56.76/-- \\
uDSR & 29.5/12.4 & 36.0/8.0 & 50.0/25.0 & 29.2/8.3 & 9.1/6.8 & -- \\
LaSR & 16.0/10.1 & 13.3/6.7 & 16.7/9.3 & 7.0/7.0 & 9.9/6.8 & 50.45/-- \\

\multicolumn{7}{@{}l}{External LLM-based (worst-5\% trimmed):} \\
SR-Scientist (GPT-OSS-120B) & 63.6/49.4 & 74.7/60.0 & \textbf{81.5}/64.8 & \textbf{66.7}/43.1 & 40.9/34.1 & -- \\
SR-Scientist (Qwen-480B) & 49.1/24.6 & 86.7/69.3 & 40.7/5.6 & 50.0/26.4 & 34.1/13.6 & -- \\
LLM-SR (Qwen-480B) & 41.1/18.1 & 80.0/52.0 & 36.1/9.3 & 30.6/18.1 & 28.8/6.1 & -- \\
A-SR (Llama3.1-8B) & 48.3/-- & 68.0/-- & 34.7/-- & 50.0/-- & 21.6/-- & --/40.5 \\

\midrule
\multicolumn{7}{@{}l}{Ours (untrimmed):} \\
\textbf{Model + tools} & \textbf{79.1}/\textbf{71.3} & \textbf{92.0}/\textbf{88.0} & 72.2/\textbf{69.4} & 58.3/\textbf{45.8} & \textbf{88.6}/\textbf{77.3} & \textbf{77.5}/\textbf{75.7} \\
\bottomrule
\end{tabular*}
\end{center}
\vspace{-2mm}
\end{table*}

Table~\ref{tab:lsr} presents the primary comparison on LSR-Synth (129 problems), where the
Qwen3.6-35B-A3B controller with the evolved library surpasses all external baselines on the
aggregate, with the widest margins on physics and at the stricter $10^{-3}$ tolerance. Because
published results use the less stringent trimmed-5\% convention, the all-points comparison is
conservative: our aggregate of 79.1 already exceeds the best external result of 63.6 (SR-Scientist), and
under the matched trimmed-5\% convention it rises to 96.1/95.4. On LSR-Transform-111, the system
reaches 77.5\% at Acc@$10^{-1}$ and 75.7\% at Acc@$10^{-2}$ (89.2/83.8 with trimmed-5\% scoring),
compared with PySR at 56.76\% (Acc@$10^{-1}$), LaSR at 50.45\% (Acc@$10^{-1}$), and A-SR at 40.5\%
(Acc@$10^{-2}$).

\begin{table*}[t]
\caption{SRSD-Feynman-240 against published systems, as accuracy (\%) at $R^2>0.999$ per difficulty
tier, on the standard and dummy-variable variants, the merged hard tiers, and the pool over all 240.
Bold marks the best value per column.}
\label{tab:other_acc}
\begin{center}
\small
\setlength{\tabcolsep}{4pt}
\begin{tabular*}{\textwidth}{@{\extracolsep{\fill}}lcccccccc@{}}
\toprule
& \multicolumn{3}{c}{\bf SRSD standard} & \multicolumn{3}{c}{\bf SRSD dummy vars.} & {\bf Hard} & {\bf Pooled} \\
\cmidrule(lr){2-4}\cmidrule(lr){5-7}
{\bf Method} & Easy & Med. & Hard & Easy & Med. & Hard & {\bf (merged)} & {\bf (240)} \\
\midrule
uDSR & \textbf{100.0} & \textbf{75.0} & 20.0 & 53.3 & 37.5 & 12.0 & 16.0 & 44.6 \\
PySR & 66.7 & 45.0 & 38.0 & 20.0 & 10.0 & 2.0 & 20.0 & 28.3 \\
DSR  & 63.3 & 45.0 & 28.0 & 76.7 & 45.0 & 22.0 & 25.0 & 42.9 \\
\midrule
\textbf{Model + tools (ours)} & 83.3 & 67.5 & \textbf{58.0} & \textbf{86.7} & \textbf{60.0} & \textbf{44.0} & \textbf{51.0} & \textbf{63.8} \\
\bottomrule
\end{tabular*}
\end{center}
\vspace{-2mm}
\end{table*}

Evaluation also moves beyond LLM-SRBench to SRSD-Feynman and CP3-28. Their formula distributions
differ from that of LLM-SRBench, so these panels test generalization across scientific equation
families (Table~\ref{tab:other_acc},
Figure~\ref{fig:cp3_ode}). On SRSD-Feynman-240, under the benchmark's original $R^2>0.999$ metric,
the system reaches $63.8$ pooled over all 240 problems against a published best of $44.6$ (uDSR), and
$58.0$ on the hard tier against $38.0$; on all dummy-variable tiers, which require rejection of
irrelevant inputs, it exceeds every published system, while uDSR
keeps the easy/medium lead (Appendix~\ref{app:srsd}). On CP3-28, where the source publishes only
per-problem MSE, we normalize each problem by the variance of its official test targets and aggregate
as the mean $\log_{10}$NMSE. The trained system obtains $-7.27$, compared with $-6.20$ for the
best published reference (QLattice). We treat this as a reference comparison. Our value uses the
official scoring files, whereas the external values are the source's order-of-magnitude scores.
Appendix~\ref{app:cp3} reports the aggregation procedure and the paired per-problem comparisons.

\subsection{Ablations}
\label{sec:exp-compose}

Table~\ref{tab:ablation} reports raw candidate quality before post-hoc selection on every benchmark;
the rows vary the two components the method evolves, first the tool library and then the model that
controls it. A reference run outside the table sets the floor: the same Qwen backbone with no tool
library at all, proposing equations directly from the data, solves 0\% of LSR-Synth at both
tolerances. The first two blocks isolate Phase~A: under the same deterministic traversal, the
initial pre-evolution library reaches 41.1/30.2 on LSR-Synth, 18.0 on SRSD hard, and $-4.76$ on
CP3-28, and the evolved library lifts these to 61.2/50.4, 29.0, and $-6.86$ with no model and no
training involved: the evolved tools carry much of the discovery capability on their own. Policy
training then adds value where tool use must be deliberate and the problems are hardest: on the
fixed evolved toolkit, the trained controller reaches 63.6/54.3 on LSR-Synth versus 54.3/46.5 for
the untrained controller, surpasses the deterministic traversal on every panel, and posts its largest
gains on the difficult problems, raising the SRSD hard score from 30.0 to 37.0 and CP3 from
$-6.69$ to $-6.99$. These pre-selection comparisons expose the quality of the candidate sets
produced by each component; final submitted scores use the common selection procedure described
in Appendix~\ref{app:selection}.

The frozen toolkit contains 33 tools organized around diagnosis, discovery, and verification;
Appendix~\ref{app:taxonomy} records its composition, the capability assets produced by Phase~A.

\begin{table*}[t]
\caption{Ablations over the two components the method evolves: the tool library (Phase~A) and the
model that controls it (Phase~B). \emph{Tools evolved} distinguishes the final evolved library
(\checkmark) from the initial library before evolution; \emph{Model used/trained} distinguish
deterministic traversal (no model), the untrained controller, and the trajectory-trained
controller. Rows without BIC report raw candidate quality before post-hoc reselection.
Slash-separated entries report the first/second threshold: LSR-Synth uses
Acc@$10^{-2}$/Acc@$10^{-3}$ and LSRT111 uses Acc@$10^{-1}$/Acc@$10^{-2}$. CP3-28 reports mean
$\log_{10}$NMSE $\downarrow$, where lower is better.}
\label{tab:ablation}
\begin{center}
\scriptsize
\setlength{\tabcolsep}{1.1pt}
\begin{tabular*}{\textwidth}{@{\extracolsep{\fill}}l@{\hspace{3pt}}ccccccccc@{}}
\toprule
& {\bf Tools} & \multicolumn{2}{c}{\bf Model} & {\bf BIC} & {\bf LSR-Synth} & {\bf LSRT111} & {\bf SRSD-240} & {\bf SRSD hard} & {\bf CP3-28} \\
{\bf Condition} & evolved & used & trained & & Acc@$10^{-2}$/Acc@$10^{-3}$ & Acc@$10^{-1}$/Acc@$10^{-2}$ & pooled & merged & $\log_{10}$NMSE$\,\downarrow$ \\
\midrule
Tools only (initial) & --   & --         & --         & --         & 41.1/30.2 & 6.3/3.6   & 29.6 & 18.0 & $-4.76$ \\
             & --          & --         & --         & \checkmark & 43.4/31.0 & 6.3/3.6   & 33.3 & 23.0 & $-5.29$ \\
\midrule

Tools only (evolved) & \checkmark & --   & --         & --         & 61.2/50.4 & 53.2/49.5 & 51.3 & 29.0 & $-6.86$ \\
             & \checkmark  & --         & --         & \checkmark & 77.5/69.8 & \textbf{77.5}/\textbf{75.7} & 62.1 & 45.0 & $-6.91$ \\
\midrule

Untrained + tools & \checkmark & \checkmark & --      & --         & 54.3/46.5 & 49.5/45.0 & 52.5 & 30.0 & $-6.69$ \\
             & \checkmark  & \checkmark & --         & \checkmark & 66.7/58.1 & 63.1/59.5        &  59.6  &  41.0  &  $-6.98$      \\
\midrule

Model + tools & \checkmark & \checkmark & \checkmark & --         & 63.6/54.3 & 53.6/50.0 & 53.3 & 37.0 & $-6.99$ \\
 & \checkmark & \checkmark & \checkmark & \checkmark & \textbf{79.1}/\textbf{71.3} & \textbf{77.5}/\textbf{75.7} & \textbf{63.8} & \textbf{51.0} & $\mathbf{-7.27}$ \\
\bottomrule
\end{tabular*}
\end{center}
\end{table*}

\section{Discussion}
\label{sec:discussion}

\paragraph{OOD behavior and formula-distribution checks.}
LLM-SRBench defines OOD by shifting the input ranges while retaining the same 129 tasks, selected
expressions, candidate pool, and custodian scorer. The controller reaches 85.3/78.3 all-points and
95.4/93.8 trimmed-5\% on this split, compared with 79.1/71.3 and 96.1/95.4 under the corresponding
in-domain conventions. The higher OOD all-points value is consistent with the relative-error score
being less sensitive to small targets after the range shift: 62\% of tasks have a larger minimum
$|y|$ on OOD, and the fraction with $\min|y|<10^{-3}$ falls from 55.0\% to 47.3\%. Formula-level
leakage is checked separately. Phase~B formulas pass the three-way overlap firewall
(Appendix~\ref{app:firewall}), and SRSD-Feynman and CP3-28 are held out from both tool evolution
and trajectory generation. Their equation families differ from LLM-SRBench, yet the system remains
strong on both panels. Together, these checks show that the capability transfers across input
ranges and formula families.

\paragraph{Dynamics of capability growth.}
The ablations above compare completed systems; here we trace how capability accumulated,
replaying the ten intermediate tool libraries recorded during Phase~A under deterministic tool
traversal;
epochs~1--9 use no BIC reselection,
\begin{wrapfigure}{r}{0.50\linewidth}
    \centering
    \includegraphics[width=0.92\linewidth]{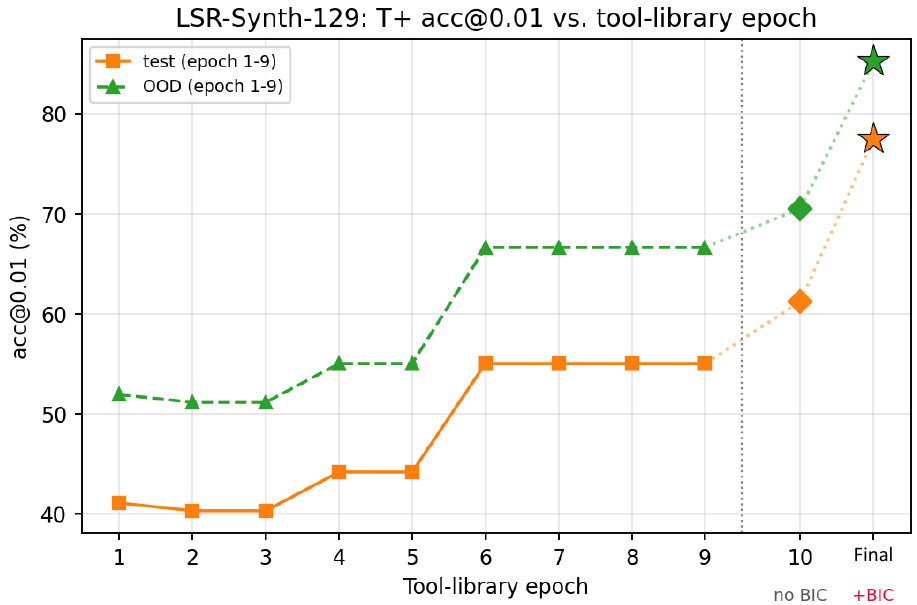}
    \caption{Capability growth on LSR-Synth-129 under deterministic tool traversal, scored with
    SR-Scientist's trimmed-5\% convention so the reference line is like-for-like. Curves show
    in-domain and shifted-input Acc@$10^{-2}$ for epochs~1--9; diamonds mark the epoch-10 no-BIC
    point and stars the BIC-selected endpoint. The top axis counts the executable analysis tools
    per snapshot. The horizontal line marks SR-Scientist (GPT-OSS-120B), which trains the model on
    a fixed toolset.}
    \label{fig:growth-curve}
\end{wrapfigure}
and epoch~10 reports both
the no-BIC anchor and the final BIC-selected point.
The resulting curve
(Figure~\ref{fig:growth-curve}) separates two mechanisms.
As the executable library grows from 23 to 32 tools (of the 25 initial tools, two are interactive
queries that traversal cannot exercise),
the no-BIC in-domain accuracy rises from
$41.1\%$ at epoch~1 to $61.2\%$ at epoch~10 under the all-points convention of
Table~\ref{tab:ablation}, and BIC reselection raises the final point to
$77.5\%$; on CP3-28 the corresponding mean $\log_{10}$NMSE moves from $-4.76$ to $-6.86$ and then
to $-6.91$.
The figure scores the same replay under SR-Scientist's trimmed-5\% convention:
SR-Scientist trains its model on a fixed toolset and reports $63.6\%$, where the initial library
already starts; evolution lifts the traversal to $86.1\%$ at epoch~10 and $95.4\%$ with BIC,
with no model in the loop. The plateaus coincide with
epochs that introduce no new effective operation on the panel, whereas the larger jumps follow
additions that widen the tested function-family set: a diagnosed missing operation changes the
library, and the enlarged library creates new candidates.

\section{Conclusion}
Scientific equation discovery is a sequential interaction between a model and numerical instruments.
An episode can fail because the library lacks an operation the formula family needs, or because the
model fails to use an operation already available. Phase~A addresses the first cause by adding
tools from diagnosed failures; Phase~B addresses the second by distilling successful tool-use
episodes into the controller. Under one verifiable feedback signal, failures expand what the system
can test and successes teach it how to search, and the coupling remains effective on shifted inputs
and on equation families outside LLM-SRBench: a recipe for scientific agents that improve their
instruments and their search behavior from the evidence of discovery.

\subsection*{AI use statement}
We used generative AI tools to assist with manuscript drafting and editing and with authoring analysis
and plotting scripts; all AI-assisted text was reviewed and revised by the authors, and all AI-assisted
code was verified against the underlying run artifacts. LLMs are additionally the object of study:
a frontier API drives the Phase~A evolution loop and an open model is trained in Phase~B, as described in
the main text. No generative AI tool was used to produce, select, or transcribe any reported number; all
numbers are copied from the official scoring files produced by the evaluation procedure of
Appendix~\ref{app:recipe}. We take
responsibility for the final content of this work, including any text, claims, or artifacts
produced with the aid of generative AI.

\subsection*{Ethics statement}
This work uses only synthetic and publicly available benchmark datasets; no human subjects or personal
data are involved. The main integrity questions it raises, benchmark leakage and memorization, are
treated as first-class subjects of the paper: Appendix~\ref{app:tool-check} examines the evolved tools
for leakage, Appendix~\ref{app:firewall} specifies the training-data overlap firewall, and
Appendix~\ref{app:recipe} discloses the accounting decisions that affect the reported numbers.

\subsection*{Reproducibility statement}
The tool-evolution loop and its interface-level data-access guarantees
(Section~\ref{sec:method-evolve}), the training-data provenance procedure, the overlap firewall,
and the information-boundary invariants (Section~\ref{sec:method-coevolve}; Appendices~\ref{app:firewall}
and~\ref{app:invariants}) are all specified in full. Appendix~\ref{app:recipe} gives the principal
training recipe, including the base model, data pipeline, LoRA settings, the shared evaluation
procedure, and the end-to-end reproduction pipeline. The evolution record
and the panel builder/scorer scripts will be released with the code.

\bibliography{references}

@article{cranmer2023pysr,
  author={Cranmer, Miles},
  title={Interpretable Machine Learning for Science with {PySR} and {SymbolicRegression.jl}},
  journal={arXiv preprint arXiv:2305.01582},
  year={2023},
  eprint={2305.01582},
  archivePrefix={arXiv},
  url={https://arxiv.org/abs/2305.01582}
}

@inproceedings{petersen2021dsr,
  author={Petersen, Brenden K. and Landajuela, Mikel and Mundhenk, T. Nathan and Santiago, Claudio P. and Kim, Soo K. and Kim, Joanne T.},
  title={Deep Symbolic Regression: Recovering Mathematical Expressions from Data via {Risk-Seeking} Policy Gradients},
  booktitle={International Conference on Learning Representations},
  year={2021},
  eprint={1912.04871},
  archivePrefix={arXiv},
  url={https://arxiv.org/abs/1912.04871}
}

@inproceedings{landajuela2022udsr,
  author={Landajuela, Mikel and Lee, Chak Shing and Yang, Jiachen and Glatt, Ruben and Santiago, Claudio P. and Mundhenk, T. Nathan and Aravena, Ignacio and Mulcahy, Garrett and Petersen, Brenden K.},
  title={A Unified Framework for Deep Symbolic Regression},
  booktitle={Advances in Neural Information Processing Systems},
  volume={35},
  pages={33985--33998},
  year={2022},
  url={https://papers.neurips.cc/paper_files/paper/2022/hash/dbca58f35bddc6e4003b2dd80e42f838-Abstract-Conference.html}
}

@inproceedings{shojaee2025llmsr,
  author={Shojaee, Parshin and Meidani, Kazem and Gupta, Shashank and Farimani, Amir Barati and Reddy, Chandan K.},
  title={{LLM-SR}: Scientific Equation Discovery via Programming with Large Language Models},
  booktitle={International Conference on Learning Representations},
  year={2025},
  eprint={2404.18400},
  archivePrefix={arXiv},
  url={https://arxiv.org/abs/2404.18400}
}

@inproceedings{grayeli2024lasr,
  author={Grayeli, Arya and Sehgal, Atharva and Costilla-Reyes, Omar and Cranmer, Miles and Chaudhuri, Swarat},
  title={Symbolic Regression with a Learned Concept Library},
  booktitle={Advances in Neural Information Processing Systems},
  volume={37},
  year={2024},
  eprint={2409.09359},
  archivePrefix={arXiv},
  url={https://arxiv.org/abs/2409.09359}
}

@article{zhao2026asr,
  author={Zhao, Wenxiao and Liu, Dong and Xu, Kaiyi and Liu, Feng and Zhao, Zhen and Ben, Fei and Wang, Shu and Li, Wenhao and Wu, Ying Nian and Ling, Fenghua and Li, Haobo and Bai, Lei},
  title={{A-SR}: {Self-Evolving} Agentic {LLMs} for Symbolic Regression via Hierarchical Coordination},
  journal={arXiv preprint arXiv:2608.04872},
  year={2026},
  eprint={2608.04872},
  archivePrefix={arXiv},
  url={https://arxiv.org/abs/2608.04872}
}

@inproceedings{xia2026srscientist,
  author={Xia, Shijie and Sun, Yuhan and Liu, Pengfei},
  title={{SR-Scientist}: Scientific Equation Discovery With Agentic {AI}},
  booktitle={International Conference on Learning Representations},
  year={2026},
  eprint={2510.11661},
  archivePrefix={arXiv},
  url={https://arxiv.org/abs/2510.11661}
}

@inproceedings{shojaee2025llmsrbench,
  author={Shojaee, Parshin and Nguyen, Ngoc-Hieu and Meidani, Kazem and Barati Farimani, Amir and Doan, Khoa D. and Reddy, Chandan K.},
  title={{LLM-SRBench}: A New Benchmark for Scientific Equation Discovery with Large Language Models},
  booktitle={Proceedings of the 42nd International Conference on Machine Learning},
  series={Proceedings of Machine Learning Research},
  volume={267},
  pages={55325--55359},
  publisher={PMLR},
  year={2025},
  eprint={2504.10415},
  archivePrefix={arXiv},
  url={https://proceedings.mlr.press/v267/shojaee25a.html}
}

@article{wang2024voyager,
  author={Wang, Guanzhi and Xie, Yuqi and Jiang, Yunfan and Mandlekar, Ajay and Xiao, Chaowei and Zhu, Yuke and Fan, Linxi and Anandkumar, Anima},
  title={Voyager: An {Open-Ended} Embodied Agent with Large Language Models},
  journal={Transactions on Machine Learning Research},
  year={2024},
  eprint={2305.16291},
  archivePrefix={arXiv},
  url={https://arxiv.org/abs/2305.16291}
}

@inproceedings{qian2023creator,
  author={Qian, Cheng and Han, Chi and Fung, Yi R. and Qin, Yujia and Liu, Zhiyuan and Ji, Heng},
  title={{CREATOR}: Tool Creation for Disentangling Abstract and Concrete Reasoning of Large Language Models},
  booktitle={Findings of the Association for Computational Linguistics: {EMNLP} 2023},
  pages={6922--6939},
  publisher={Association for Computational Linguistics},
  year={2023},
  eprint={2305.14318},
  archivePrefix={arXiv},
  doi={10.18653/v1/2023.findings-emnlp.462},
  url={https://aclanthology.org/2023.findings-emnlp.462/}
}

@inproceedings{yuan2024craft,
  author={Yuan, Lifan and Chen, Yangyi and Wang, Xingyao and Fung, Yi R. and Peng, Hao and Ji, Heng},
  title={{CRAFT}: Customizing {LLMs} by Creating and Retrieving from Specialized Toolsets},
  booktitle={International Conference on Learning Representations},
  year={2024},
  eprint={2309.17428},
  archivePrefix={arXiv},
  url={https://openreview.net/forum?id=G0vdDSt9XM}
}

@inproceedings{wang2024trove,
  author={Wang, Zhiruo and Neubig, Graham and Fried, Daniel},
  title={{TroVE}: Inducing Verifiable and Efficient Toolboxes for Solving Programmatic Tasks},
  booktitle={Proceedings of the 41st International Conference on Machine Learning},
  series={Proceedings of Machine Learning Research},
  volume={235},
  pages={51177--51191},
  publisher={PMLR},
  year={2024},
  eprint={2401.12869},
  archivePrefix={arXiv},
  url={https://proceedings.mlr.press/v235/wang24az.html}
}

@inproceedings{hu2025adas,
  author={Hu, Shengran and Lu, Cong and Clune, Jeff},
  title={Automated Design of Agentic Systems},
  booktitle={International Conference on Learning Representations},
  year={2025},
  eprint={2408.08435},
  archivePrefix={arXiv},
  url={https://proceedings.iclr.cc/paper_files/paper/2025/hash/36b7acf6f6010652b3f2a433774a66fe-Abstract-Conference.html}
}

@article{novikov2025alphaevolve,
  author={Novikov, Alexander and Vu, Ngan and Eisenberger, Marvin and Dupont, Emilien and Huang, Po-Sen and Wagner, Adam Zsolt and Shirobokov, Sergey and Kozlovskii, Borislav and Ruiz, Francisco J. R. and Mehrabian, Abbas and Kumar, M. Pawan and See, Abigail and Chaudhuri, Swarat and Holland, George and Davies, Alex and Nowozin, Sebastian and Kohli, Pushmeet and Balog, Matej},
  title={{AlphaEvolve}: A Coding Agent for Scientific and Algorithmic Discovery},
  journal={arXiv preprint arXiv:2506.13131},
  year={2025},
  eprint={2506.13131},
  archivePrefix={arXiv},
  url={https://arxiv.org/abs/2506.13131}
}

@inproceedings{schick2023toolformer,
  author={Schick, Timo and Dwivedi-Yu, Jane and Dessi, Roberto and Raileanu, Roberta and Lomeli, Maria and Hambro, Eric and Zettlemoyer, Luke and Cancedda, Nicola and Scialom, Thomas},
  title={Toolformer: Language Models Can Teach Themselves to Use Tools},
  booktitle={Advances in Neural Information Processing Systems},
  volume={36},
  year={2023},
  eprint={2302.04761},
  archivePrefix={arXiv},
  url={https://proceedings.neurips.cc/paper/2023/hash/d842425e4bf79ba039352da0f658a906-Abstract-Conference.html}
}

@inproceedings{patil2024gorilla,
  author={Patil, Shishir G. and Zhang, Tianjun and Wang, Xin and Gonzalez, Joseph E.},
  title={Gorilla: Large Language Model Connected with Massive {APIs}},
  booktitle={Advances in Neural Information Processing Systems},
  volume={37},
  year={2024},
  eprint={2305.15334},
  archivePrefix={arXiv},
  doi={10.52202/079017-4020},
  url={https://proceedings.neurips.cc/paper_files/paper/2024/hash/e4c61f578ff07830f5c37378dd3ecb0d-Abstract-Conference.html}
}

@inproceedings{qin2024toollm,
  author={Qin, Yujia and Liang, Shihao and Ye, Yining and Zhu, Kunlun and Yan, Lan and Lu, Yaxi and Lin, Yankai and Cong, Xin and Tang, Xiangru and Qian, Bill and Zhao, Sihan and Hong, Lauren and Tian, Runchu and Xie, Ruobing and Zhou, Jie and Gerstein, Mark and Li, Dahai and Liu, Zhiyuan and Sun, Maosong},
  title={{ToolLLM}: Facilitating Large Language Models to Master 16000+ Real-world {APIs}},
  booktitle={International Conference on Learning Representations},
  year={2024},
  eprint={2307.16789},
  archivePrefix={arXiv},
  url={https://arxiv.org/abs/2307.16789}
}

@article{qu2024toolsurvey,
  author={Qu, Changle and Dai, Sunhao and Wei, Xiaochi and Cai, Hengyi and Wang, Shuaiqiang and Yin, Dawei and Xu, Jun and Wen, Ji-Rong},
  title={Tool Learning with Large Language Models: A Survey},
  journal={arXiv preprint arXiv:2405.17935},
  year={2024},
  eprint={2405.17935},
  archivePrefix={arXiv},
  url={https://arxiv.org/abs/2405.17935}
}

@inproceedings{qian2025toolrl,
  author={Qian, Cheng and Acikgoz, Emre Can and He, Qi and Wang, Hongru and Chen, Xiusi and Hakkani-Tur, Dilek and Tur, Gokhan and Ji, Heng},
  title={{ToolRL}: Reward is All Tool Learning Needs},
  booktitle={Advances in Neural Information Processing Systems},
  volume={38},
  year={2025},
  eprint={2504.13958},
  archivePrefix={arXiv},
  doi={10.52202/085713-3524},
  url={https://arxiv.org/abs/2504.13958}
}

@article{brunton2016sindy,
  author={Brunton, Steven L. and Proctor, Joshua L. and Kutz, J. Nathan},
  title={Discovering Governing Equations from Data by Sparse Identification of Nonlinear Dynamical Systems},
  journal={Proceedings of the National Academy of Sciences},
  volume={113},
  number={15},
  pages={3932--3937},
  year={2016},
  eprint={1509.03580},
  archivePrefix={arXiv},
  doi={10.1073/pnas.1517384113},
  url={https://doi.org/10.1073/pnas.1517384113}
}

@inproceedings{cai2024latm,
  author={Cai, Tianle and Wang, Xuezhi and Ma, Tengyu and Chen, Xinyun and Zhou, Denny},
  title={Large Language Models as Tool Makers},
  booktitle={International Conference on Learning Representations},
  year={2024},
  eprint={2305.17126},
  archivePrefix={arXiv},
  url={https://arxiv.org/abs/2305.17126}
}

@article{du2024llmed,
  author={Du, Mengge and Chen, Yuntian and Wang, Zhongzheng and Nie, Longfeng and Zhang, Dongxiao},
  title={Large Language Models for Automatic Equation Discovery of Nonlinear Dynamics},
  journal={Physics of Fluids},
  volume={36},
  number={9},
  pages={097121},
  year={2024},
  eprint={2405.07761},
  archivePrefix={arXiv},
  doi={10.1063/5.0224297},
  url={https://arxiv.org/abs/2405.07761}
}

@inproceedings{merler2024icsr,
  author={Merler, Matteo and Haitsiukevich, Katsiaryna and Dainese, Nicola and Marttinen, Pekka},
  title={{In-Context} Symbolic Regression: Leveraging Large Language Models for Function Discovery},
  booktitle={Proceedings of the 62nd Annual Meeting of the Association for Computational Linguistics {(Volume} 4: Student Research Workshop)},
  pages={427--444},
  publisher={Association for Computational Linguistics},
  year={2024},
  eprint={2404.19094},
  archivePrefix={arXiv},
  doi={10.18653/v1/2024.acl-srw.49},
  url={https://aclanthology.org/2024.acl-srw.49/}
}

@inproceedings{behzadifar2026decaevolve,
  author={Behzadifar, Pouya and Shojaee, Parshin and Kabra, Sanchit and Meidani, Kazem and Reddy, Chandan K.},
  title={{DecAEvolve}: Decompose, Adapt, and Evolve, or, Three Pillars of Effective {LLM-based} Scientific Equation Discovery},
  booktitle={Proceedings of the 43rd International Conference on Machine Learning},
  series={Proceedings of Machine Learning Research},
  volume={306},
  address={Seoul, South Korea},
  publisher={PMLR},
  year={2026},
  url={https://radical-p.github.io/DecAEvolve/}
}

@article{huang2026skillselfplay,
  author={Huang, Siyuan and Cheng, Pengyu and Liu, Haotian and Chen, Tao and Liu, Yihao and Ni, Jingwei and Zhou, Shijie and Yang, Ziyi and Jiang, Gangwei and Zhou, Mengyu and Cheng, Yu and Jiang, Xiaoxi and Jiang, Guanjun},
  title={Skill {Self-Play}: Pushing the Frontier of {LLM} Capability with {Co-Evolving} Skills},
  journal={arXiv preprint arXiv:2607.22529},
  year={2026},
  eprint={2607.22529},
  archivePrefix={arXiv},
  url={https://arxiv.org/abs/2607.22529}
}

@inproceedings{cheng2026mem2evolve,
  author={Cheng, Zihao and Liu, Zeming and Shan, Yingyu and Wang, Xinyi and Zhu, Xiangrong and Ma, Yunpu and Wang, Hongru and Guo, Yuhang and Lin, Wei and Wang, Yunhong},
  title={{Mem2Evolve}: Towards {Self-Evolving} Agents via {Co-Evolutionary} Capability Expansion and Experience Distillation},
  booktitle={Proceedings of the 64th Annual Meeting of the Association for Computational Linguistics {(Volume} 1: Long Papers)},
  pages={20784--20831},
  publisher={Association for Computational Linguistics},
  year={2026},
  eprint={2604.10923},
  archivePrefix={arXiv},
  url={https://aclanthology.org/2026.acl-long.952/}
}

@article{hu2026seal,
  author={Hu, Yihao and Wen, Zhihao and Liu, Xiujin and Wang, Pan and Zhang, Xin and Wu, Wei},
  title={{SEAL}: Synergistic {Co-Evolution} of Agents and Learning Environments},
  journal={arXiv preprint arXiv:2605.24426},
  year={2026},
  eprint={2605.24426},
  archivePrefix={arXiv},
  url={https://arxiv.org/abs/2605.24426}
}

@article{chen2026evotrainer,
  author={Chen, Guhong and Shi, Yingcheng and Li, Yongbin and Li, Binhua and Xu, Xander and Wei, Hu and Ni, Shiwen and Yang, Min and Ye, Jieping},
  title={{EvoTrainer}: {Co-Evolving} {LLM} Policies and Training Harnesses for Autonomous Agentic Reinforcement Learning},
  journal={arXiv preprint arXiv:2606.03108},
  year={2026},
  eprint={2606.03108},
  archivePrefix={arXiv},
  url={https://arxiv.org/abs/2606.03108}
}

@inproceedings{fang2025webevolver,
  author={Fang, Tianqing and Zhang, Hongming and Zhang, Zhisong and Ma, Kaixin and Yu, Wenhao and Mi, Haitao and Yu, Dong},
  title={{WebEvolver}: Enhancing Web Agent {Self-Improvement} with Co-evolving World Model},
  booktitle={Proceedings of the 2025 Conference on Empirical Methods in Natural Language Processing},
  pages={8959--8975},
  publisher={Association for Computational Linguistics},
  year={2025},
  eprint={2504.21024},
  archivePrefix={arXiv},
  doi={10.18653/v1/2025.emnlp-main.454},
  url={https://aclanthology.org/2025.emnlp-main.454/}
}

@inproceedings{yang2026coevolve,
  author={Yang, Shidong and Ma, Ziyu and Huang, Tongwen and Hu, Yiming and Wang, Yong and Chu, Xiangxiang},
  title={{CoEvolve}: Training {LLM} Agents via {Agent-Data} Mutual Evolution},
  booktitle={Proceedings of the 64th Annual Meeting of the Association for Computational Linguistics {(Volume} 1: Long Papers)},
  pages={23015--23036},
  publisher={Association for Computational Linguistics},
  year={2026},
  eprint={2604.15840},
  archivePrefix={arXiv},
  doi={10.18653/v1/2026.acl-long.1055},
  url={https://aclanthology.org/2026.acl-long.1055/}
}

@inproceedings{yang2026evotool,
  author={Yang, Shuo and Han, Caren and Ma, Xueqi and Li, Yan and Ghasemi Madani, Mohammad Reza and Hovy, Eduard},
  title={{EVOTOOL}: {Self-Evolving} {Tool-Use} Policy Optimization in {LLM} Agents via {Blame-Aware} Mutation and {Diversity-Aware} Selection},
  booktitle={Proceedings of the 64th Annual Meeting of the Association for Computational Linguistics {(Volume} 1: Long Papers)},
  pages={43553--43572},
  publisher={Association for Computational Linguistics},
  year={2026},
  doi={10.18653/v1/2026.acl-long.2016},
  url={https://aclanthology.org/2026.acl-long.2016/}
}

@inproceedings{wolflein2025toolmaker,
  author={Wolflein, Georg and Ferber, Dyke and Truhn, Daniel and Arandjelovic, Ognjen and Kather, Jakob Nikolas},
  title={{LLM} Agents Making Agent Tools},
  booktitle={Proceedings of the 63rd Annual Meeting of the Association for Computational Linguistics {(Volume} 1: Long Papers)},
  pages={26092--26130},
  publisher={Association for Computational Linguistics},
  year={2025},
  doi={10.18653/v1/2025.acl-long.1266},
  url={https://aclanthology.org/2025.acl-long.1266/}
}

@article{yang2026kepleragent,
  author={Yang, Jianke and Venkatachalam, Ohm and Kianezhad, Mohammad and Vadgama, Sharvaree and Yu, Rose},
  title={Think like a Scientist: Physics-guided {LLM} Agent for Equation Discovery},
  journal={arXiv preprint arXiv:2602.12259},
  year={2026},
  eprint={2602.12259},
  archivePrefix={arXiv},
  url={https://arxiv.org/abs/2602.12259}
}

@article{matsubara2022srsd,
  author={Matsubara, Yoshitomo and Chiba, Naoya and Igarashi, Ryo and Ushiku, Yoshitaka},
  title={Rethinking Symbolic Regression Datasets and Benchmarks for Scientific Discovery},
  journal={Journal of Data-centric Machine Learning Research (DMLR)},
  year={2024},
  eprint={2206.10540},
  archivePrefix={arXiv},
  url={https://arxiv.org/abs/2206.10540}
}

@misc{openai2026gpt55,
  author={OpenAI},
  title={Introducing GPT-5.5},
  year={2026},
  month={apr},
  url={https://openai.com/index/introducing-gpt-5-5/},
  note={Product release and model documentation}
}

@misc{thing2024cp3bench,
  author={Thing, Mattias E. and Koksbang, Sofie M.},
  title={cp3-bench: A Tool for Benchmarking Symbolic Regression Algorithms Tested with Cosmology},
  year={2024},
  eprint={2406.15531},
  archivePrefix={arXiv},
  url={https://arxiv.org/abs/2406.15531}
}
\bibliographystyle{iclr2027_conference}

\appendix
\section{Appendix}

\subsection{CP3-28: error-magnitude aggregate and per-method head-to-head}
\label{app:cp3}
\paragraph{Mean $\log_{10}$NMSE (Figure~\ref{fig:cp3_ode}).} The figure summarizes the overall
CP3 comparison. The benchmark publishes per-problem MSEs but no common aggregate. We normalize each
problem as $\text{NMSE}=\text{MSE}/
\mathrm{Var}(y_{\text{test}})$ with the variance taken over the official 2{,}000-row test targets,
and report the mean $\log_{10}$NMSE over the 28 problems.
Our values use the official scoring files; published per-problem values use the source's reported
order-of-magnitude scores.

\paragraph{Per-problem pairing (Table~\ref{tab:cp3}).} The table complements the figure with a
per-problem win/tie/loss breakdown against the 12 published methods for which the source provides
paired values. Ties use the source's order-of-magnitude resolution. After the common candidate-selection
pass, the aggregate values for the trained controller with the evolved library and for deterministic
traversal are $-7.27$ and $-6.91$, respectively.

\begin{table}[h]
\caption{CP3-28 (cosmology) against all 12 published methods. First column: the method's mean
$\log_{10}$NMSE $\downarrow$ (procedure above; lower is better; after candidate selection, our reference values are $-7.27$ for
the trained controller with the library and $-6.91$ for deterministic traversal).
Remaining columns: per-problem head-to-head as win\,\%\,/\,tie\,\%\,/\,loss\,\% of paired problems,
ordered by increasing trained-controller win rate. The pairing is at order-of-magnitude grain (equal
$\lfloor\log_{10}\text{MSE}\rfloor$ counts as a tie), which is conservative for us; denominators vary
($22$--$28$) because the source omits some per-problem cells.}
\label{tab:cp3}
\begin{center}
\footnotesize
\setlength{\tabcolsep}{6pt}
\begin{tabular}{@{}lccc@{}}
\toprule
{\bf vs.} & {\bf mean $\log_{10}$NMSE $\downarrow$} & {\bf Tools only} & {\bf Model + tools} \\
\midrule
QLattice & $-6.20$ & 30.8 / 34.6 / 34.6 & 34.6 / 42.3 / 23.1 \\
ITEA     & $-6.16$ & 25.9 / 40.7 / 33.3 & 44.4 / 25.9 / 29.6 \\
GPG      & $-4.45$ & 40.7 / 40.7 / 18.5 & 48.1 / 37.0 / 14.8 \\
GPZGD    & $-4.59$ & 60.7 / 25.0 / 14.3 & 64.3 / 21.4 / 14.3 \\
Operon   & $-2.66$ & 64.3 / 28.6 / \phantom{0}7.1 & 67.9 / 25.0 / \phantom{0}7.1 \\
GE       & $-3.66$ & 70.4 / 25.9 / \phantom{0}3.7 & 70.4 / 25.9 / \phantom{0}3.7 \\
uDSR     & $-5.02$ & 62.5 / 20.8 / 16.7 & 70.8 / 12.5 / 16.7 \\
PySR     & $-5.09$ & 72.7 / 22.7 / \phantom{0}4.5 & 77.3 / 18.2 / \phantom{0}4.5 \\
FFX      & $-3.84$ & 81.5 / 14.8 / \phantom{0}3.7 & 81.5 / 14.8 / \phantom{0}3.7 \\
AI-Feynman & $-4.16$ & 84.0 / 12.0 / \phantom{0}4.0 & 88.0 / \phantom{0}8.0 / \phantom{0}4.0 \\
DSO      & $-3.05$ & 88.5 / \phantom{0}7.7 / \phantom{0}3.8 & 88.5 / \phantom{0}7.7 / \phantom{0}3.8 \\
DSR      & $-3.16$ & 88.9 / \phantom{0}7.4 / \phantom{0}3.7 & 88.9 / \phantom{0}7.4 / \phantom{0}3.7 \\
\bottomrule
\end{tabular}
\end{center}
\end{table}

\subsection{SRSD-Feynman evaluation under the original benchmark metric}
\label{app:srsd}
Published SRSD-Feynman systems report accuracy at $R^2>0.999$ per difficulty tier (easy/medium/hard, 30/40/50
problems), with a parallel set of dummy-variable tiers in which irrelevant input columns are added
and a solution counts only if it ignores them \citep{matsubara2022srsd}. Table~\ref{tab:other_acc}
reports this metric per tier and pooled over all 240 problems; the external rows are the published
per-tier values of \citet{matsubara2022srsd} (Tables~4--5), and their pooled column is computed from
those values with the same 30/40/50 weights. On our side, per-problem $R^2$
values come from the official scorer, with the common final-selection procedure described in
Appendix~\ref{app:selection}. The training corpus of Section~\ref{sec:method-coevolve} has verified
zero overlap with the SRSD formulas.

On the hard tier, model~+~tools reaches $58.0\%$ against a published best of $38.0\%$ (PySR), and
$51.0\%$ on the merged hard tiers against a published best of $25.0\%$ (DSR). On all three
dummy-variable tiers, our conditions exceed every published system, while uDSR remains ahead on the
easy and medium standard tiers ($100.0/75.0$). The largest gains therefore occur on harder tiers and
on tasks that require rejecting irrelevant variables.

\subsection{The self-evolution loop in full}
\label{app:loop}
Algorithm~\ref{alg:evolve} states the Phase~A loop of Section~\ref{sec:method-evolve} precisely.
For readability the algorithm tracks the tool library $\mathcal{L}$ separately and writes
$\mathcal{H}$ for the remaining harness components (the interaction interface $\rho$ and selection
rule $\sigma$). The state carried across batches is $\mathcal{L}$, $\mathcal{H}$, and the set
$\mathcal{S}$ of problems solved so far. The acceptance check on line~\ref{line:gate} re-runs the
accumulated suite before committing an edit, preserving previously solved capability. A newly authored
tool also passes the held-out promotion gate on line~\ref{line:promote} before entering the persistent
library.

\begin{algorithm}[h]
\caption{Failure-driven harness evolution with diagnosis and monotone acceptance.}
\label{alg:evolve}
\begin{algorithmic}[1]
\State \textbf{Input:} problem stream partitioned into batches $B_1, B_2, \dots$; acting model
$M$; reflection model $R$
\State \textbf{Initialize:} tool library $\mathcal{L}\gets\mathcal{L}_0$ (generic seed set); harness $\mathcal{H}\gets
\mathcal{H}_0$; solved set $\mathcal{S}\gets\emptyset$
\For{each batch $B_t$ in the stream}
  \For{each problem $p\in B_t$}
    \State $(u, \hat c) \gets \textsc{Solve}(M, p, \mathcal{L}, \mathcal{H})$ \Comment{trajectory $u$, proposed equation $\hat c$}
    \If{$\textsc{Verify}(\hat c, p)$ passes}
      \State $\mathcal{S}\gets\mathcal{S}\cup\{p\}$
    \Else
      \State $a \gets \textsc{Diagnose}(R, u, p)\in\{\textsc{Capability},\textsc{Selection}\}$
      \If{$a=\textsc{Capability}$}
        \State $g \gets \textsc{BuildOrRepairTool}(R, u)$
        \If{$\textsc{PromotionGate}(g)$ passes} \label{line:promote}
          \State $\mathcal{L}' \gets \mathcal{L}\cup\{g\}$
        \Else{} \textbf{continue}
        \EndIf
        \State $\mathcal{H}' \gets \mathcal{H}$
      \Else
        \State $\mathcal{H}' \gets \textsc{EditHarness}(R, u)$
        \State $\mathcal{L}' \gets \mathcal{L}$
      \EndIf
      \If{$\textsc{AcceptanceCheck}(\mathcal{L}', \mathcal{H}', \mathcal{S})$} \label{line:gate}
        \State $\mathcal{L}\gets\mathcal{L}'$;\quad $\mathcal{H}\gets\mathcal{H}'$
      \EndIf
    \EndIf
  \EndFor
\EndFor
\State \textbf{return} frozen $\mathcal{L}$ and $\mathcal{H}$
\end{algorithmic}
\end{algorithm}

\subsection{The frozen toolkit by role}
\label{app:taxonomy}
This appendix describes the frozen library tool by tool. Each tool carries an explicit
role assigned at creation, and the roles cover the canonical steps of the empirical loop: look at the
data, hypothesize a form, fit it, and check it. The frozen library holds 33 tools: 10 diagnostics, 18
discovery fitters, 4 verifiers, and the restricted Python builder.

Diagnostics read the data and report structure without committing to a form: detecting periodicity by
spectral analysis, estimating local power-law exponents from log--log slopes, probing variable
dependence through numerical derivatives, searching for separability and normalized interactions, and
exposing radial/angular structure. Discovery, the largest group, proposes and fits forms:
general-purpose fitters (sparse feature regression over polynomial, trigonometric, exponential, and
interaction features; additive-monomial, multiplicative-separable, and separable nonlinear
least-squares fitters) and residual-boosting tools, together with a few mechanism fitters that carry a
domain prior in the shapes they stage (examined in Appendix~\ref{app:tool-check}). Verification scores and
stress-tests candidates, fitting on held data, estimating a posterior over a component library, and
probing stability under bootstrap and holdout perturbations. The builder is the restricted-Python tool
builder of Section~\ref{sec:method-evolve}, the mechanism by which the loop extends its own capability;
it is frozen along with the rest during Phase~B.

\subsection{Leakage checks}
\label{app:tool-check}
The library is developed on the LLM-SRBench Phase-A stream. SRSD-Feynman and CP3-28 are external
panels not used in tool evolution or model training. Together with the Phase-B overlap firewall,
this analysis examines whether
the reusable procedure carries across panels without benchmark-specific formulas.

\paragraph{(1) Domain-flavored mechanism fitters encode scientific priors.}
A few discovery tools stage textbook families such as strain polynomials, Arrhenius terms,
temperature peaks, and saturating or oscillatory kinetics. Their coefficients are fitted from public
rows, and the tools receive neither benchmark metadata nor target expressions. The staged families
therefore serve as data-tested hypotheses; held-out and OOD scoring determines whether a family
generalizes beyond the development stream.

\paragraph{(2) The numeric fitters enumerate generic shapes programmatically.}
General-purpose fitters enumerate variable-based fragments such as $\sqrt{v}$, $\log v$,
$\exp(-v)$, $\sin(kv)$, products, ratios, and radial/angular combinations. Free parameters are
fitted from the data, and the candidate-generation path contains no response-column lookup table or
benchmark-specific constants.

\paragraph{(3) Ground-truth candidate pools are isolated to oracle probes.}
The task representation also supports an optional candidate pool containing a reference expression
and decoys for internal oracle probes. Only the oracle skill and oracle archive read this pool;
benchmark loaders initialize it empty, and the primary generator receives no pool or metadata. These
probes measure tool sufficiency and decoy robustness and are reported separately from benchmark scores.

\subsection{What did not work}
\label{app:failures}
Three extensions of the coupled procedure were evaluated. Large-scale reinforcement learning did not
improve on the supervised trajectory recipe, so Phase~B uses the simpler positive-signal distillation
setup. Online tool authoring likewise produced no measurable gain; tool creation remains in the
offline evolution phase while solving focuses on selecting and composing the frozen library. Finally,
the trained controller did not replace the frontier model in Phase~A. The resulting system is
asymmetric: a stronger model grows the harness, and a smaller open model learns to operate it.

\subsection{The overlap firewall}
\label{app:firewall}
This section formalizes the disjointness guarantee used for the training-data of
Section~\ref{sec:method-coevolve}. Let $\mathcal{E}$ be the set of all published evaluation formulas
(the union of every benchmark panel with a public ground-truth expression). A synthesized candidate
formula $c$ is admitted to the training pool only if it collides with no $e\in\mathcal{E}$
under any of three relations; Algorithm~\ref{alg:firewall} states the test.

\paragraph{(i) Canonical symbolic equality.}
Let $\kappa(\cdot)$ map an expression to a canonical symbolic normal form (a deterministic
simplification and normalization of the expression tree, including variable renaming to a canonical
order). Candidate $c$ collides with $e$ if $\kappa(c)=\kappa(e)$, i.e., $c$ and $e$ are the same
expression up to writing.

\paragraph{(ii) Affine equivalence.}
Two formulas are treated as the same problem if one is an affine reparametrization of the other,
since such a rescaling is trivially absorbed by fitting. Formally, $c$ collides with $e$ if there
exist an input affine map $\phi(\vx)=\mathrm{diag}(\va)\vx+\vb$ and output scalars $\alpha\neq 0,
\beta$ such that $\kappa\big(\alpha\,c(\phi(\vx))+\beta\big)=\kappa(e(\vx))$.

\paragraph{(iii) Numeric fingerprint.}
As a safety net against normalization gaps, $c$ and $e$ are also compared numerically: on a fixed
bank of shared probe points $\{\vx^{(k)}\}_{k=1}^{K}$ drawn once and reused, $c$ collides with $e$
if the (scale-normalized) values agree to tolerance $\varepsilon$, i.e.
$\max_k \big| \tilde c(\vx^{(k)}) - \tilde e(\vx^{(k)})\big| < \varepsilon$, where $\tilde{\,\cdot\,}$
denotes standardization that quotients out an overall affine factor. A candidate that agrees
pointwise with an evaluation formula is rejected even if its symbolic form was written differently.

\begin{algorithm}[h]
\caption{Overlap firewall: admit a synthesized candidate only if disjoint from all evaluation formulas.}
\label{alg:firewall}
\begin{algorithmic}[1]
\State \textbf{Input:} candidate $c$; evaluation formula set $\mathcal{E}$; probe bank
$\{\vx^{(k)}\}_{k=1}^{K}$; tolerance $\varepsilon$
\For{each $e\in\mathcal{E}$}
  \If{$\kappa(c)=\kappa(e)$} \Return \textsc{Reject} \Comment{(i) canonical symbolic}
  \EndIf
  \If{$\exists\,(\va,\vb,\alpha,\beta):\ \kappa(\alpha\,c(\phi(\vx))+\beta)=\kappa(e)$}
    \Return \textsc{Reject} \Comment{(ii) affine equivalence}
  \EndIf
  \If{$\max_k |\tilde c(\vx^{(k)})-\tilde e(\vx^{(k)})| < \varepsilon$}
    \Return \textsc{Reject} \Comment{(iii) numeric fingerprint}
  \EndIf
\EndFor
\State \Return \textsc{Admit} \Comment{disjoint under all three relations}
\end{algorithmic}
\end{algorithm}

Panels without a public ground-truth formula contribute no members to $\mathcal{E}$ and, separately,
are excluded from training, so overlap with them cannot arise. The full record of admit/reject
decisions is retained and released with the code, together with the information-boundary invariants
enforced at trajectory materialization (Section~\ref{sec:method-coevolve}).

\subsection{Information-boundary invariants}
\label{app:invariants}
Every training example is derived from a recorded trajectory that satisfies the following information
boundaries:
\begin{enumerate}
  \item Episode completeness: the trajectory terminates through the normal procedure.
  \item No held-out rows read: no operation in the episode accessed the evaluation split.
  \item No selection labels leaked: selection-set target values never appeared in any
  message visible to the model.
  \item No ground-truth read: the custodian manifest holding target formulas was never
  accessed by the model or the trainer.
  \item No private-reward feedback: the scalar reward used for bookkeeping was never
  returned into the model's context.
\end{enumerate}

\subsection{Candidate-selection procedure}
\label{app:selection}
This appendix records the details of the candidate-selection procedure of
Section~\ref{sec:method-coevolve}. One selection rule is fixed per metric family, chosen once and
applied symmetrically to the deterministic-traversal and trained-controller conditions (never tuned
per panel): the BIC rule of Equation~\ref{eq:bic} on the panels scored by accuracy at a tolerance
(LSR-Synth and LSRT111), and the error on the selection subset with a light complexity term on the
regression-quality panels (SRSD-Feynman and CP3-28). Every selection rule reads only public
observations: in Equation~\ref{eq:bic} the $n$ public observations pool the fitting and selection
subsets together, and no test observation enters any selection rule on any panel. For the
trained-controller condition, a model-proposed candidate is promoted over the deterministic tool
anchor only when it improves by a fixed margin on both public subsets, namely the error on the
selection subset and the error on observations withheld within the fitting subset; otherwise the
tool anchor is kept.

\subsection{Training recipe}
\label{app:recipe}
This appendix summarizes the training and evaluation recipe for the reported Qwen3.6-35B-A3B
controller, its SFT adapter, and the evolved library.

\paragraph{Base model and tools.}
The controller is Qwen3.6-35B-A3B (35B total and approximately 3B active parameters), run in
bfloat16. The same 33-tool catalog is frozen for training and evaluation. The comparison conditions
are tools-only deterministic traversal, an untrained controller with the library, and the
trajectory-trained controller with the library.

\paragraph{Training data.}
Training uses 72 programmatically generated SR tasks instantiated in the evaluation-time format. Their
reference expressions pass the three-way overlap firewall against the published evaluation formulas
(Appendix~\ref{app:firewall}). Teacher trajectories are rolled by GPT-5.5 with the frozen harness, and retained
examples contain positive-signal actions with their associated result context. The corpus is split by
task group for training and validation.

\paragraph{SFT hyperparameters.}
The adapter is trained with token-level cross-entropy on the retained action tokens. We use LoRA
(rank 16, $\alpha=32$) with frozen base weights, a constant learning rate of
$1\!\times\!10^{-6}$, and 120 optimization steps; the resulting adapter is approximately 42\,MB.

\paragraph{Shared evaluation procedure.}
All conditions run over the frozen catalog under one shared promotion rule. The model receives feedback computed from
the fitting subset; the selection subset is used for final candidate choice, while the test and OOD
data are scored offline. BIC reselection on the Acc@$\tau$ panels is a post-hoc pass over public
observations only (Appendix~\ref{app:selection}).

\paragraph{Reproduction pipeline.}
The release scripts generate the overlap-checked synthetic panel, roll and materialize trajectories,
train the adapter, and re-evaluate all panels under the fixed procedure.

\paragraph{Accounting.}
Infrastructure failures count as misses and do not reduce the reported denominators. The tool-free
GPT-5.5 reference in Figure~\ref{fig:cp3_ode} receives the public observations and numeric feedback
without the local fitting library; it is included as a contextual reference.

\end{document}